# Anthropomorphic Features for On-Line Signatures

Moises Diaz, Miguel A. Ferrer, and Jose J. Quintana

*Abstract*—Many features have been proposed in on-line signature verification. Generally, these features rely on the position of the online signature samples and their dynamic properties, as recorded by a tablet. This paper proposes a novel feature space to describe efficiently on-line signatures. Since producing a signature requires a skeletal arm system and its associated muscles, the new feature space is based on characterizing the movement of the shoulder, the elbow and the wrist joints when signing. As this motion is not directly obtained from a digital tablet, the new features are calculated by means of a virtual skeletal arm (VSA) model, which simulates the architecture of a real arm and forearm. Specifically, the VSA motion is described by its 3D joint position and its joint angles. These anthropomorphic features are worked out from both pen position and orientation through the VSA forward and direct kinematic model. The anthropomorphic features' robustness is proved by achieving state-of-the-art performance with several verifiers and multiple benchmarks on third party signature databases, which were collected with different devices and in different languages and scripts.

## 1. INTRODUCTION

The handwritten signature is used world-wide for authentication, possibly because of its non-invasive characteristics. Countries with different lifestyles, cultures, religious or political systems use signatures constantly. Examples of the high acceptability of its use are found in signing various agreements, marriage registers or generic contracts, for last wills and testaments, bank checks, autographs and so on. The physical movement associated with signature execution gives an insight into the neuromotor state of the signer and his/her motor system. As such, the motor skills are analyzed through dynamic signatures in kinesiology, psychiatry, neurology, education or, in general, e-health applications. The key to these analyses is the huge growth in signature acquisition technology, especially those related to improvements in sensor quality. There is also greater use of signature verification in security environments, communications, in modern e-government applications, questioned document examination and other biometric systems. In the case of biometrics [1], signatures are popular for access control, person verification, financial transactions and so on. Indeed, signatures can offer greater security and convenience than other methods which are token-based (e.g., passports or ID cards) or knowledge-based (e.g., PINs or passwords).

There are two major challenges in signature-based biometric systems: the unpredictable intra- and inter- personal variability. The former refers to the differences between signatures from the same signer and the latter to strategies for faking the identity of signers. In this context, systems have to deal with two kinds of forgeries. On the one hand, random forgeries (RFs), which try to verify the identity of a signer by using their own genuine signatures. On the other hand, skilled forgeries (SFs) reproduce the genuine signature of a signer after the forger learns by practice the intrapersonal variability of the victim.

### 1.1. Related Work

In biometrics, automatic signature verifiers (ASVs) are designed to deal with off-line or on-line signatures. According to this nomenclature, an offline signature refers to a scanned image which represents an inked specimen whereas on-line signatures represent signatures executed over a digital tablet. On-line ASVs are generally the most accurate because the signatures contain the temporal and dynamic order in which the specimens were executed [2].

As a two-class problem, an ASV typically has to decide if a questioned signature belongs to the claimed identity or not. A large number of systems in the literature respond to this challenge. They are commonly based on original feature extraction followed by the use of competitive classifiers, as seen in the review [3], [4]. The present work is focused on on-line signature modality and a novel proposal for extracting anthropomorphic features, which are deduced from pen-tip movement. As such, this section covers related works in features extracted in online signature verification. Digital tablets commonly provide the position and timing function features as well as the pressure and pen-tip angles.

This information can be used in on-line ASVs as global or local features. Global features refer to single parameters that describe the whole signature (i.e., duration, average velocity, number of pen downs, etc.). Local features refer to the information extracted at a particular point in the signature, for example, the velocity, acceleration or pressure [3], [4].

The most popular type of on-line automatic signature verifier is based on Dynamic Time Warping (DTW). These systems develop a Euclidean [5], [6] or City block [7] matrix between the timing function features provided from the digital tablet, mainly the trajectory coordinates and pressure [6], [8]. Other systems augment these signals with their first and second derivatives [9], [10]. The dissimilarity between the two signatures is worked out by looking for the minimum cost path through the matrix. Some corpuses also provide the pen-tip azimuth and inclination [11].

There are other proposals which use mathematical transformations of the time-based functions in order to add further information to the matrix. Some examples are the log curvature ratios, stroke length to width ratios or simply the sine and cosine values [12]. The feature set is often reduced by using some feature selection algorithm [2]. For instance, a proposal in [13] suggests including only the more stable features among the signatures in the reference set. Other authors have found it more convenient to use only the information contained in the x and y coordinates [14].

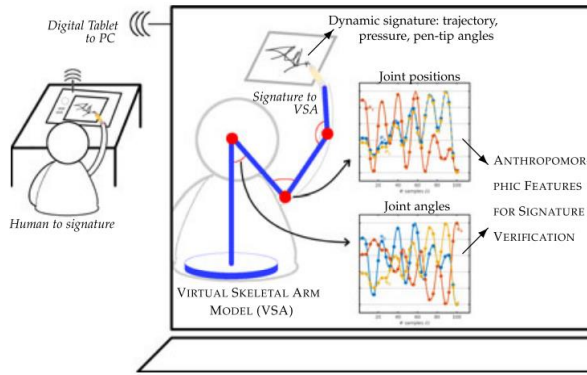

Fig. 1. Anthropomorphic features for signature verification systems, extracted by a Virtual Skeletal Arm model.

## 1.2. Our Contribution

The innovative contribution of this paper is the potential use of anthropomorphic features in signature verification. If we are able to reproduce the 3D joints movement from a 2D signature trajectory, a more discriminative feature space can be expected for signature verification. We suggest its achievement by means of a Virtual Skeletal Arm (VSA) model, based on the architecture of an anthropomorphic robotic arm, as illustrated in Fig. 1.

Because of the motion of the VSA model, a set of anthropomorphic features is obtained from changes in both the joint angles and joint positions in the upper limb. In order to achieve this, a set of non-linear transformations is carried out to calculate the anthropomorphic features, all of which have a physical meaning.

Estimating the arm posture from the trajectory of the signature is a problem with a number of solutions. However, we can reduce the range of possibilities by fixing an ergonomic posture of the VSA model when signing. Thus, we provide a simple, fast to compute and verifiable solution. Furthermore, competitive results are achieved by using the anthropomorphic features, without the need to fine-tune, them in a comprehensive evaluation. All in all, this paper makes the following main contributions.

1) New anthropomorphic features, extracted from a novel feature space, based on the arm's posture when signing. These features consist of the sequence of 3D joint positions of the arm and the corresponding sequence of joint angles.

2) The mathematical fundamentals for applying VSA models to signature verification by simulating the architecture of a real arm and forearm. The anthropomorphic features are obtained through the forward and direct kinematic model of the VSA.

3) A thorough study of the initial set up of the VSA to produce efficient anthropomorphic features.

4) An adaptation of two different state-of-the art verifiers and a study of the fusion of two types of anthropomorphic features at feature and score level.

5) A series of challenging trials with multiple databases and verifiers. We use third party databases and verifiers, which are not employed at the setting up stage.

In reporting this research, we share the open source code to carry out the space transformation from pen-tip orientation and position to the anthropomorphic features through the VSA model. Therefore, another novel contribution of this work is the anthropomorphic feature extractor, which can be downloaded from www.gpds.ulpgc.es.

This paper is organized as follows. In Section 2, we present our virtual skeletal arm model, while the mathematical formulation to calculate the anthropomorphic features is detailed in Section 3. In Section 4, two verifiers are adapted to use the anthropomorphic features. An optimal configuration of the initial posture of the VSA to improve performance is given in Section 5 and experimental results are provided in Section 6. Final remarks are given in Section 7.

## 2. VIRTUAL SKELETAL ARM MODEL

The theoretical arm skeletal system [15] comprises bones, joints and muscles. These anatomical structures have to work in coordination to produce motion such as handwriting. The bones and cartilage are the rigid parts which shape and support the whole arm [15]. One of the regions of the appendicular skeleton comprises the bones of the upper limbs, which are formed by the humerus, ulna, radius and the hand bones. The humerus is the only bone in the upper arm. The shoulder joint provides articulation between the glenoid cavity of the scapula (shoulder blade) and the head of the humerus. The other end of the bone is articulated with the ulna and the radius bones forming the elbow. Regarding the forearm, the ulna is longer than the radius. Both articulate with the carpal bones, forming the wrist joint. The hand consists of the carpus, the metacarpus and the phalanges of the fingers. The metacarpus is composed of five long bones called metacarpals. They are articulated with the bones of the fingers, which are two for the thumb and three for the rest of the fingers.

The arm joints [15] are the connections between the rigid components of the bones. These joints allow motion of the arm. The upper limbs can be described as follows [15]. The shoulder links the arm to the trunk. It is the most flexible joint in the human body, allowing free movement of the arm. In an idle state, the arm hangs vertically beside the trunk. The elbow is the central joint of the upper limbs. Similar to a hinge, the elbow can be bent in a range of angles. The intercarpal joints gather the carpal joints of the wrist. These joints, when considered together, provide movement in all possible directions. Additionally, the finger bones can move because of the interphalangeal joints, located between the phalanges of the hand. The motion of this system is generated by the contraction and relaxation of the muscles.

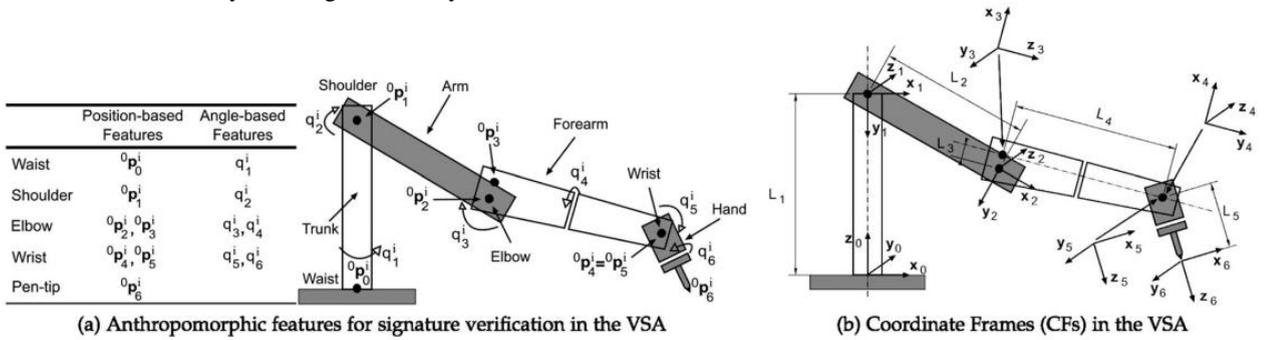

(a) Anthropomorphic features for signature verification in the VSA     (b) Coordinate Frames (CFs) in the VSA

Fig. 2. Architecture of the Virtual Skeletal Arm (VSA) model.

In order to obtain the anthropomorphic features, a Virtual Skeletal Arm model is developed, which is based on the structure of the anthropomorphic robotic arm ABB IRB 120. This robot has a considerable similarity to the upper extremities since its architecture represents the waist, trunk, shoulder, arm and forearm as well as a simplified mechanism for holding a pen. The objective in modeling such a robotic arm is to be able to verify that the model arm can effectively reproduce the given signature using the estimated anthropomorphic features as input.

The links of the VSA are connected in a way similar to realistic bone connections. The trunk link is connected to the base of our VSA model, $L_1$ being its length, as illustrated in Fig. 2b. This is directly connected to the arm or humerus-based link, where $L_2$ is its length. Its other side is connected to the forearm or ulna and radius-based link, of length $L_4$. Because of the architecture of the commercial arm-robot we used, the union between $L_2$ and $L_4$ is slightly biased. This leads to these axes not being fully aligned and this, therefore, generates a corresponding constant distance $L_3$. This misalignment will require some angle corrections in the ensuing model. The forearm, or ulna and radius-based link, is connected to the final link, which includes the pen, $L_5$ being its length.

The motion of the VSA is constrained by the arm's movements. The position of the waist is indicated at instant $i$ by ${}^0P_0^i$. It has a rotation $q_1^i$, which is produced along the vertical (or yaw) axis through the trunk. In the shoulder joint, ${}^0P_1^i$. being its position, a flex extension $q_2^i$ is allowed along the roll axis through the union of the trunk and arm or humerus-based link. In the case of the elbow, there are two equivalent positions with the same motion because of $L_3$, i.e., ${}^0P_2^i$ and ${}^0P_3^i$. In the elbow joint, a rotation along the roll axis $q_3^i$ is allowed, which covers the full angular range of a human elbow, which is about $(0.1\pi\text{-}\pi)$. The elbow also allows a rotation, which affects the ulna and radius as well as the wrist and hand. This rotation is approximated by the angle $q_4^i$, along the pitch axis. The wrist position is characterized by ${}^0P_4^i$. and ${}^0P_5^i$. since it has two rotations in the VSA: a rotation along the roll axis $q_5^i$ and a rotation along the pitch axis $q_6^i$. The latter simulates the angular motion of the simplified hand-holding, where its position is determined by ${}^0P_6^i$. These angles and positions can be identified in Fig. 2a, which determines the anthropomorphic features.

Thus, the anthropomorphic features can be represented by i ) seven joint positions referred to the base of the VSA, called position-based anthropomorphic features P(${}^0P_k^i$), $\forall k \in \{0, ... ,6\}$ and, ii) six joint angles, called angle-based anthropomorphic features Q ($q_k^i$), $\forall k \in \{0, ... ,6\}$. These values depend on the pen-tip orientation and ballistic trajectory coordinates of the signature, sampled with $\forall k \in \{0, .. ,m\}$., m being the total number of samples, and they depend on the initial position of the VSA.

The mathematical formulation of the anthropomorphic features is given in the next section. However, for research purposes, we share the open source code from www.gpds. ulpgc.esto extract the proposed features.

## 3. ANTHROPOMORPHIC FEATURE EXTRACTION

To work out the anthropomorphic features, we need to define the Coordinate Frames (CF) of the VSA's joints. A Coordinate Frame of a joint contains a reference system with the joint as the coordinate origin. Specifically, the CF in the base, or waist, of the VSA model is designated by {$S_0$}. The CF in the shoulder is characterized by{$S_1$}. The elbow joint has two CFs because of the specific VSA model design: they are {$S_2$} and {$S_3$}. This procedure allows us to convert the biased angles of this particular modelled robot to unbiased elbow angles. Since the wrist has two movements, it is represented with two CFs: {$S_4$} and {$S_5$}. Finally, the CF {$S_6$} is assigned to the pen-tip. Fig. 2b illustrates the CFs assigned to each joint. The mathematical notations used in this section are shown in Table 1.is assigned to the pen-tip. Fig. 2b illustrates the CFs assigned to each joint. The mathematical notations used in this section are shown in Table 1.

Once the CFs are defined, the transformation matrices between the different coordinate frames [16] are then calculated. The matrix ${}^0T_6^i$ defines the pose of the system {$S_6$} taking as its reference the system{$S_0$}, and changes along the signature

$$
{}^0T_6^i = \begin{pmatrix} n_x^i & o_x^i & a_x^i & p_x^i \\ n_y^i & o_y^i & a_y^i & p_y^i \\ n_z^i & o_z^i & a_y^i & p_z^i \\ 0 & 0 & 0 & 1 \end{pmatrix}
\tag{1}
$$

TABLE 1
Notation Used in the Anthropomorphic Feature Extraction

| Notation | Description |
|---|---|
| VSA | Virtual Skeletal Arm |
| ${}^0\mathbf{p}_k^i$ | 3D position of the joint $k$ referred to {$S_0$} for a sampling point $i$ |
| $\mathbf{P}({}^0\mathbf{p}_k^i)$ | Position-based anthropomorphic features |
| $q_k^i$ | Joint angle $k$ for the sampling point $i$ |
| $\mathbf{Q}(q_k^i)$ | Angle-based anthropomorphic features |
| ${}^a\mathbf{T}_b^i$ | Homogeneous transformation matrix related to reference systems {$S_a$} and {$S_b$} for a sampling point $i$ in the signature |
| {$S_k$} | Coordinate frame $k$ |
| $s^i$ | On-line signature with $m$ sampling points |
| $\theta^i, \phi^i$ | Azimuth and inclination angles |
| $[\delta_k^i, d_k, a_k, \alpha_k]$ | Denavit Hartenberg parameters ($\mathbf{DH}_k^i$) |
| $k$ | Index for the anthropomorphic features |
| $i$ | Index for on-line signature sampling points |
| $\omega$ | weighting factor for score fusion |
| $\Gamma = (\rho_x, \rho_y, \rho_z)$ | 3D rotation of the writing planes |

According to the first elements of its columns, the vectors ${}^0n_6^i = (n_x^i, n_y^i, n_z^i)^T$, $o_6^i = (o_x^i, o_y^i, o_z^i)^T$ and $a_6^i = (a_x^i, a_y^i, a_z^i)^T$are respectively the direction vectors of the axes $x_6$, $y_6$, $z_6$. The vector ${}^0p_6^i = (p_x^i, p_y^i, p_z^i)^T$ then indicates the position of {$S_6$}. All of these vectors are referred to CF {$S_0$}. Finally, the last row will always have the values shown in Eq. (1) for the further homogeneous transformation matrices used in this work.

Since {$S_6$}is over the pen-tip, the pose of the signature is assigned to it. An on-line signature is described by the pen-position, pen-azimuth and pen-inclination respectively as $s^i = (p_x^i, p_y^i, p_z^i, \theta^i, \emptyset^i)$. It is typically acquired from a digital device which has its own CF. Let {$S_s$}be the coordinate frame of a signature, located in a corner of the writing surface. In Fig. 3 the spatial relationship between CF{$S_0$}, {$S_6$}and {$S_s$} is shown along with its associated homogeneous transformation matrices

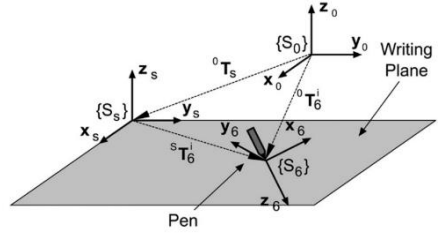

Fig. 3. Coordinate frames $\{S_0\}$, $\{S_6\}$ and $\{S_s\}$ and their associated homogeneous transformation matrices $^0\mathrm{T}_6^i$, $^0\mathrm{T}_s$ and $^s\mathrm{T}_6^i$.

Because the homogeneous transformation matrix can be composed, it can be written $^0 T_6^i = \, ^0\mathrm{T}_S \cdot \, ^0 T_6^i$ Where, $^0\mathrm{T}_S$ is a constant matrix and $^0 T_6^i$ depends on the pen-tip position. The sequence of matrices $^0 T_6^i$ can be calculated by means of a translation and two rotations.The translation is determined by the position of the pen-tip, as related to the CF$\{S_s\}$. On the first rotation, $\mathrm{Rot}(\, z_s, \theta^i) = \, ^{\ell 1} T_{\ell_2}^i$, $z_6$ and $z_s$ are aligned when the CF $\{S_s\}$ is rotated $(\frac{\pi}{2} - \theta^i)$ on the $z_s$ axis.

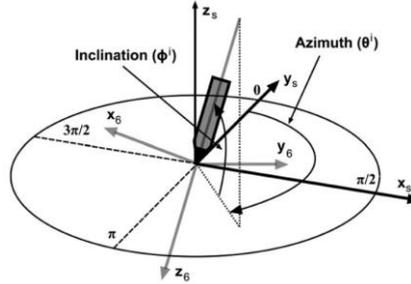

Fig. 4. Pen orientation $\{S_0\}$ with respect to the writing surface $\{S_s\}$.

On the second rotation, $\mathrm{Rot}(y_s, \emptyset^i) = \, ^{\ell 2} T_6^i$ is rotated $(-\frac{\pi}{2} - \emptyset^i)$ on the new $y_s$ axis. Such geometric operations can be deduced from Fig. 4, whereas the matrices that represent these two rotations are shown below, where the indexes $\ell_1$ and $\ell_2$ represent intermediate transformations

$$^s T_{\ell 1}^i = \begin{pmatrix} 1 & 0 & 0 & p_x^i \\ 0 & 1 & 0 & p_y^i \\ 0 & 0 & 1 & p_z^i \\ 0 & 0 & 0 & 1 \end{pmatrix} \tag{2}$$

$$^{\ell 1} T_{\ell_2}^i = \begin{pmatrix} \cos(\frac{\pi}{2} - \theta^i) & -\sin(\frac{\pi}{2} - \theta^i) & 0 & 0 \\ \sin(\frac{\pi}{2} - \theta^i) & \cos\left(\frac{\pi}{2} - \theta^i\right) & 0 & 0 \\ 0 & 0 & 1 & 0 \\ 0 & 0 & 0 & 1 \end{pmatrix} \tag{3}$$

$$^{\ell 2} T_6^i = \begin{pmatrix} \cos\left(\frac{\pi}{2} + \emptyset^i\right) & 0 & -\sin\left(\frac{\pi}{2} + \emptyset^i\right) & 0 \\ 0 & 1 & 0 & 0 \\ \sin\left(\frac{\pi}{2} + \emptyset^i\right) & 0 & \cos\left(\frac{\pi}{2} + \emptyset^i\right) & 0 \\ 0 & 0 & 0 & 1 \end{pmatrix} \tag{4}$$

By combining these movements, we have $^s T_6^i = \, ^sT_{\ell_2} \cdot \, ^{\ell 2} T_6^i$; simplifying and putting c =cos and s= sin ,we obtain

$$^s T_6^i = \begin{pmatrix} -\mathrm{s}(\theta^i)s(\emptyset^i) & -c(\theta^i) & -\mathrm{s}(\theta^i)s(\emptyset^i) & p_x^i \\ -\mathrm{c}(\theta^i)s(\emptyset^i) & s(\theta^i) & -c(\theta^i)c(\emptyset^i) & p_y^i \\ \mathrm{c}(\emptyset^i) & 0 & -s(\emptyset^i) & p_z^i \\ 0 & 0 & 0 & 1 \end{pmatrix} \tag{5}$$

The position of the {$S_s$} with respect to the base of the VSA model {$S_0$} is constant and given by $s^i = (p_x^i, p_y^i, p_z^i, \theta^i, \emptyset^i)$. . Their orientations coincide, as shown in Fig. 3. Thus, the matrix ${}^0T_S$ that relates to them is that relating to them is:

$$
{}^0T_S = \begin{pmatrix} 1 & 0 & 0 & p_{s,x} \\ 0 & 1 & 0 & p_{s,y} \\ 0 & 0 & 1 & p_{s,z} \\ 0 & 0 & 0 & 1 \end{pmatrix}
\tag{6}
$$

The relationship of a sampling point i in the on-line signature with respect to {$S_0$} can be obtained as

$$
{}^0T_6^i = {}^0T_S \cdot {}^0T_6^i = \begin{pmatrix} -s(\theta^i)s(\emptyset^i) & -c(\theta^i) & -s(\theta^i)s(\emptyset^i) & p_{s,x} + p_x^i \\ -c(\theta^i)s(\emptyset^i) & s(\theta^i) & -c(\theta^i)c(\emptyset^i) & p_{s,y} + p_y^i \\ c(\emptyset^i) & 0 & -s(\emptyset^i) & p_{s,z} + p_z^i \\ 0 & 0 & 0 & 1 \end{pmatrix}
\tag{7}
$$

### TABLE 2
### DH Parameters, $\mathbf{DH}_k^i$

| Joint $k$ | $\delta_k^i$ | $d_k$ | $a_k$ | $\alpha_k$ |
|-----------|--------------|-------|-------|------------|
| 1 | $q_1^i$ | $L_1$ | 0 | $-\frac{\pi}{2}$ |
| 2 | $q_2^i - \frac{\pi}{2}$ | 0 | $L_2$ | 0 |
| 3 | $q_3^i$ | 0 | $L_3$ | $-\frac{\pi}{2}$ |
| 4 | $q_4^i$ | $L_4$ | 0 | $\frac{\pi}{2}$ |
| 5 | $q_5^i$ | 0 | 0 | $-\frac{\pi}{2}$ |
| 6 | $q_6^i$ | $L_5$ | 0 | 0 |

Once the sequence of matrixes ${}^0T_6^i$ are computed, the anthropomorphic features can be determined. For this purpose, the following two issues need to be taken into account [16]. 1) Inverse kinematic: the formulation of the joint angle-based anthropomorphic features $Q(q_k^i)$ ,given the CFs. 2) Forward kinematic: the formulation of the position-based anthropomorphic features P (${}^0p_k^i$), given the joint angles. Although the angles are actually calculated first, for clarity purposes, we first introduce the forward kinematics.

### 3.1 Forward Kinematics of the VSA Model

The purpose of the forward kinematics is to calculate the pose of the coordinate frames (CFs) relating to the VSA model, as a function of its joints angles $Q(q_k^i)$. For this purpose, the Denavit-Hartenberg (DH) [17] algorithm is widely used. Briefly, the steps of this algorithm are described as follows:

1) As shown in Fig. 2c, a reference position is required for the VSA in order to set the relationships of CFs according to the DH rules.
2) Each joint, k, is characterized by four transformation parameters over the CF {$S_{k-1}$} to make a match to{$S_k$}: $\delta_k^i$ is the angle about $z_{k-1}$ that makes $x_{k-1}$and $x_k$ parallel. The di k angle is obtained from the angle based anthropomorphic features $Q(q_k^i)$, according to the configuration described in Fig. 2c. $d_k$ represents the offset along $z_{k-1}$ to align $x_{k-1}$ and $x_k$. $a_k$ denotes the offset along $x_k$ that puts {$S_{k-1}$} and {$S_k$}: in the same position. $a_k$ is the angle about $x_k$ that makes a match between {$S_{k-1}$} and {$S_k$}. For the VSA model, these parameters, $DH_k^i$),, are provided in Table 2.
3) The row associated with the joint k contains the geometrical information needed to build the homogeneous transformation matrix ${}^{k-1}T_k^i$, that relates the coordinate frame {$S_k$}to its immediate precedent {$S_{k-1}$}

$$^{k-1}T_k^i = \begin{pmatrix} c(\delta_k^i) & -c(\alpha_k)s(\delta_k^i) & s(\alpha_k)s(\delta_k^i) & a_k c(\delta_k^i) \\ s(\delta_k^i) & c(\alpha_k)c(\delta_k^i) & -s(\alpha_k)c(\delta_k^i) & a_k s(\delta_k^i) \\ 0 & -s(\alpha_k) & c(\alpha_k) & d_k \\ 0 & 0 & 0 & 1 \end{pmatrix}$$ (8)

The homogeneous matrix that relates the final CF to the base of the VSA model is obtained using the matrices $^{k-1}T_k^i$, as follows:

$$^0T_6^i = {}^0T_1^i \cdot {}^1T_2^i \cdot {}^2T_3^i \cdot {}^3T_4^i \cdot {}^4T_5^i \cdot {}^5T_6^i$$ (9)

Since the position $^0p_k^i$, can be collected from last column of $^0T_k^i$, we can now calculate the position of all joints by processing Eq. (9). Algorithm 1 determines the position-based anthropomorphic features, referred to the base of the VSA model.

---

**Algorithm 1.** Position $\mathbf{P}(^0\mathbf{p}_k^i)$ of the VSA Model

---

**Function:** $\mathbf{P}(^0\mathbf{p}_k^i) \leftarrow \text{GetPos}(\mathbf{DH}_k^i)$

1: **for** $k \leftarrow 1$ **to** $k \leftarrow 6$ **do**
2:     $[\delta_k^i, d_k, a_k, \alpha_k] \leftarrow \mathbf{DH}_k^i$                        ▷ DH parameters
3:     $^{k-1}\mathbf{T}_k^i \leftarrow f(\delta_k^i, d_k, a_k, \alpha_k)$            ▷ from Eq. (8)
4: **end for**
5: **for** $k \leftarrow 0$ **to** $k \leftarrow 6$ **do**
6:     **if** $k == 0$ **then**
7:         $^0\mathbf{p}_k^i \leftarrow \{0, 0, 0\}$
8:     **else if** $k > 0$ **then**
9:         $^0\mathbf{T}_k^i \leftarrow \prod_{\ell=1}^{k} {}^{\ell-1}\mathbf{T}_\ell^i$
10:       $^0\mathbf{p}_k^i \leftarrow \{^0\mathbf{T}_k^i(1,4), {}^0\mathbf{T}_k^i(2,4), {}^0\mathbf{T}_k^i(3,4)\}$
11:     **end if**
12: **end for**

---

### 3.2. Inverse Kinematics of the VSA Model

The objective of the inverse kinematics is to deduce the joint angle-based anthropomorphic features, Q $(q_k^i)$, of the VSA model, based on the pose of the pen attached to the end of the model. As shown in Fig. 2, there are six joint angles to calculate. For their resolution, kinematic decoupling is used, which divides the problem into two parts. First, the angles related to the trunk, shoulder and elbow, i.e., $q_1^i; q_2^i; q_3^i$, are calculated. They give the position $^0p_5^i$. Second, we deduce the angles related to the ulna and radius rotation $q_4^i$, the wrist $q_5^i$ and the hand-holding $q_6^i$, which provide the desired orientation.

1) The first joint angles, $q_1^i; q_2^i; q_3^i$: To solve for the first three angles, it is necessary to know the wrist position $^0p_5^i$. For this purpose, we can assume that the vector $^0p_6^i$ is obtained by summing the vectors $^0p_5^i$ and $L_5 \cdot {}^0a_6^i$ [18]. These vectors can be extracted from the matrix $^0T_6^i$ in Eq. (7). Thus, 0pi 5 can be calculated as

$$^0p_5^i = {}^0p_6^i - L_5 \cdot {}^0a_6^i$$ (10)

Once $^0p_5^i$ is determined, we can obtain the angle qi 1 by the following trigonometrical expression

$$q_1^i = \text{atan2}(p_y^i, p_x^i), \qquad (p_y^i, p_x^i) \in {}^0p_5^i$$ (11)

where the operator atan2 returns the four-quadrant arctangent of $p_y^i/p_x^i$.

To deduce $q_2^i$ and $q_3^i$, the geometrical relationships of the anthropomorphic configuration of the VSA model is shown in Fig. 5. Thus, the projection of the point $^0 p_5^i$ in the XY-plane can be written as

$$r_1^i = \sqrt{(p_x^i)^2 + p_y^i)^2} \; ; \qquad r_2^i = p_z^i - L_1; \qquad (p_y^i, p_x^i, p_z^i) \in {}^0 p_5^i \qquad (12)$$

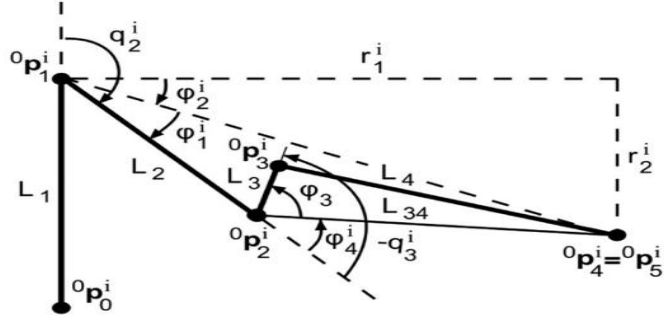

Fig. 5. Configuration of VSA for computing $q_2^i$ and $q_3^i$.

Moreover, we can define $\varphi_3 =$ atan2 $(L_4, L_3)$, $L_{34} = \sqrt{(L_3^2 + (L_4)^2}$ and $\varphi_3 + \varphi_4^i = -q_3^i$, . The angle $\varphi_4^i$ can be also obtained by the cosine law, as follows:

$$L_2^2 + L_{34}^2 + 2L_2 L_{34} \cos(r_1^i) = (r_1^i)^2 + (r_2^i)^2 \qquad (13)$$

$$\varphi_4^i = a\cos(\frac{(r_1^i)^2 + (r_2^i)^2}{L_2^2 + L_{34}^2 + 2L_2 L_{34}}) \qquad (14)$$

Therefore, we can now calculate the value of $q_3^i$ as follows:

$$q_3^i = -\varphi_3 - \varphi_4^i \qquad (15)$$

The values of $\varphi_1^i$ and $\varphi_2^i$ from Fig. 5 can be worked out through the following trigonometrical identities

$$\varphi_1^i = atan2(L_{34} \cos(\varphi_4^i), ( L_2 + L_{34} \cos(\varphi_4^i)) \qquad (16)$$

$$\varphi_2^i = atan2(r_2^i, r_1^i) \qquad (17)$$

Similarly, $q_2^i$ is obtained as

$$q_2^i = \frac{\pi}{2} + \varphi_1^i + \varphi_2^i \qquad (18)$$

At this point it should be noted that the angle of rotation of the elbow is characterized by the angle $\varphi_4^i$, so that when this angle is zero the elbow is fully extended and when this angle is close to $\pi$, it is fully contracted.

2) The last joint angles, $q_4^i$, $q_5^i$, $q_6^i$: Because of their property of combination, we can write $^0 T_6^i$ as

$$^0 T_6^i = {}^0 T_3^i \cdot {}^3 T_6^i \qquad (19)$$

Thus, the matrix $^0 T_6^i$ gives the desired pose between the base of the VSA model and the pen-tip $\{S_6\}$, as calculated in Eq. (7). Then, because of the fact that $q_1^i$, $q_2^i$ and $q_3^i$ are already known, it is possible to calculate $^0 T_3^i$, using Eq. (8). So, putting in the right side the known matrices, we can deduce

$$^3 T_6^i = ( {}^0 T_3^i )^{-1} \cdot {}^0 T_6^i \qquad (20)$$

In addition, we can apply the DH parameter to the last three joints using Eq. (8), to obtain $^3T_4^i$, $^4T_5^i$ and $^5T_6^i$ in order to determine the following mathematical relationship, where $S_k = \sin(q_k^i)$ and $C_k = \cos(q_k^i)$

$$^3T_6^i = {}^3T_4^i \cdot {}^4T_5^i \cdot {}^5T_6^i = \begin{pmatrix} c_4c_5c_6 - s_4s_6 & -c_4c_5s_6 - s_4c_6 & -c_4s_5 & -L_5c_4s_5 \\ s_4c_5c_6 + c_4s_6 & -s_4c_5s_6 + c_4c_6 & -s_4s_5 & -L_5s_4s_5 \\ s_5c_6 & -s_5s_6 & c_5 & L_4 + L_5c_5 \\ 0 & 0 & 0 & 1 \end{pmatrix} \tag{21}$$

Finally, twelve algebraic equations are obtained by equaling the terms of (20) and (21). In order to obtain the angles described in the four quadrants, we select the following equations:

$$\frac{s_4s_5}{c_4s_5} = \frac{\varphi_y^i}{\varphi_x^i} \rightarrow q_4^i = \text{atan2}(a_y^i, a_x^i) \tag{22}$$

$$-s_4s_5 = a_y^i \rightarrow q_5^i = \text{asin}(a_y^i / s_4) \tag{23}$$

$$\frac{-s_5s_6}{c_4s_5} = \frac{o_z^i}{n_z^i} \rightarrow q_6^i = \text{atan2}(-o_z^i, n_z^i) \tag{24}$$

$$(a_x^i, a_y^i, o_z^i, n_z^i) \in [({}^0T_3^i)^{-1} \cdot {}^0T_6^i]$$

Despite $q_5^i$ being limited to movement in the range $(-\frac{\pi}{2}, \frac{\pi}{2})$ it is worth pointing out that it reproduces the wrist movement in this range. In Algorithm 2 we formalize the calculation of the inverse kinematics with the VSA model for a particular on-line signature.

---

**Algorithm 2. Angle-Based Anthropomorphic Features $\mathbf{Q}(q_k^i)$**

**Function:** $\mathbf{Q}(q_k^i) \leftarrow \text{GetAng}(s^i, \mathbf{DH}_k^i)$

1: $^s\mathbf{T}_6^i \leftarrow f(p_x^i, p_y^i, p_z^i, \theta^i, \phi^i)$      ▷ from Eq. (5)
2: $^0\mathbf{T}_6^i \leftarrow {}^0\mathbf{T}_s \cdot {}^s\mathbf{T}_6^i$
3: $^0\mathbf{p}_5^i \leftarrow f({}^0\mathbf{T}_6^i, L_5)$      ▷ kinematic decoupling, Eq. (10)
4: $q_1^i \leftarrow \text{atan2}(p_y^i, p_x^i);$    $(p_x^i, p_y^i) \in {}^0\mathbf{p}_5^i$
5:      ▷ Let $(\varphi_1^i, \varphi_2^i, \varphi_3, \varphi_4^i)$ be geometrical parameters obtained from Fig. 5
6: $q_2^i \leftarrow \pi/2 + \varphi_1^i + \varphi_2^i$
7: $q_3^i \leftarrow -\varphi_3 - \varphi_4^i$
8: **for** $k \leftarrow 1$ **to** $k \leftarrow 3$ **do**
9:      $[\delta_k^i, d_k, a_k, \alpha_k] \leftarrow \mathbf{DH}_k^i$      ▷ DH parameters
10:      $^{k-1}\mathbf{T}_k^i \leftarrow f(\delta_k^i, d_k, a_k, \alpha_k)$      ▷ from Eq. (8)
11: **end for**
12: $^0\mathbf{T}_3^i \leftarrow \prod_{k=1}^{3} {}^{k-1}\mathbf{T}_k^i$
13:
14: $^3\mathbf{T}_6^i \leftarrow ({}^0\mathbf{T}_3^i)^{-1} \cdot {}^0\mathbf{T}_6^i$
15:      ▷ Let $^3\mathbf{n}_6^i, {}^3\mathbf{o}_6^i, {}^3\mathbf{a}_6^i$ be three direction vectors of $^3\mathbf{T}_6^i$
16: $q_4^i \leftarrow \text{atan2}(a_y^i, a_x^i);$    $(a_x^i, a_y^i) \in [({}^0\mathbf{T}_3^i)^{-1} \cdot {}^0\mathbf{T}_6^i]$
17: $q_5^i \leftarrow \text{asin}(-a_y^i / \sin(q_4^i));$    $a_y^i \in [({}^0\mathbf{T}_3^i)^{-1} \cdot {}^0\mathbf{T}_6^i]$
18: $q_6^i \leftarrow \text{atan2}(-o_z^i, n_z^i);$    $(o_z^i, n_z^i) \in [({}^0\mathbf{T}_3^i)^{-1} \cdot {}^0\mathbf{T}_6^i]$
19: $\mathbf{Q}(q_k^i) \leftarrow (q_1^i, q_2^i, q_3^i, q_4^i, q_5^i, q_6^i)$

---

### 3.3. VSA Kinematic Validation

As we have based our VSA model on the industrial IRB 120 arm-robot, we need to validate the mathematical equations behind the anthropomorphic features. For this purpose, three human signatures were registered in a digital tablet, the Wacom Intuos PRO. Then, through inverse kinematics, we calculated the angle-based anthropomorphic features $Q(q_k^i)$, which were used as the input to the robot. Later, the signatures were executed by the robot and registered again in the same digital tablet. In order to validate the kinematics of our VSA model, the similarities between the real and the robotic online signatures were quantified by the signal-tonoise ratio, SNR, as follows:

$$\text{SNR} = 10 \log \left( \frac{\sum_{i=1}^{n}((x_{sp}^i)^2 + (y_{sp}^i)^2)}{\sum_{i=1}^{n}((x_{sr}^i)^2 + (y_{sr}^i)^2)} \right) \tag{25}$$

Where $x_{sp}^i = (x_s^i - \overline{x_s^i})$, $y_{sp}^i = (y_s^i - \overline{y_s^i})$, $x_{sr} = (x_s^i - x_r^i)$, $y_{sr} = (y_s^i - y_r^i)$. ($x_r^i$, $y_r^i$) are the human and robotic trajectories. Each of the pieces of handwriting were interpolated in order to ensure the same number of sampling points n, and, therefore, to make the comparison with the SNR. The SNR values quantify the quality of the reconstruction between the two trajectories. In average terms, we obtained a SNR above 15 dB. For a qualitative validation, Fig. 6 exemplifies the similarities between both kinds of trajectories. We observed that the robotic signatures do not produce linear trajectories between sampling points, which slightly distorted the straight lines in the signatures. However, both numerical and visual validation confirmed to us the proposed mathematical implementation of the anthropomorphic features

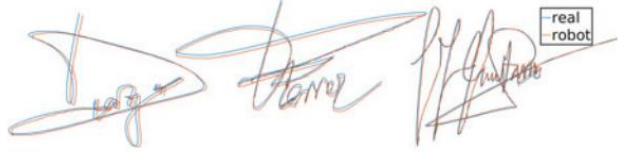

Fig. 6. Similarities in the real and robotic on-line signature trajectories.

## 4. ANTHROPOMORPHIC FEATURES IN AUTOMATIC SIGNATURE VERIFICATION

The anthropomorphic features obtained in Section 3 consist of the six joint angles $Q(q_k^i)$ and the seven joint positions $P(p_k^i)$. Regarding the positions, we can observe in Fig. 2 that $^0p_0^i$ and $^0p_1^i$ do not change for any sampling point i in the signature, $^0p_2^i$ and $^0p_3^i$ have the same movement and $^0p_4^i$ is equal to $^0p_5^i$. Therefore, the seven joint positions can be reduced to three: $^0p_2^i$ , $^0p_5^i$ and $^0p_6^i$ which refers to elbow, wrist and finger positions, respectively. Therefore, $^0p_2^i = {}^0p_e^i = (x_e^i, y_e^i, z_e^i)$ , $^0p_5^i = {}^0p_w^i = (x_w^i, y_w^i, z_w^i)$ and $^0p_6^i = {}^0p_f^i = (x_f^i, y_f^i, z_f^i)$.

In respect to the automatic signature verifier (ASV) configuration, we use two different state-of-the-art ASVs. The aim is to test the versatility of the use of anthropomorphic features in two completely different ASVs, which are each based on different methodologies. This makes them suitable for covering a wide range of signature properties, as needed for a thorough assessment of the anthropomorphic features in terms of performance achieved.

Verifier 1). Function-based + DTW. Enrolled and questioned signatures are compared by using the same DTW configuration proposed in [6]. The feature vector can be built using position-based features, $V_p = (x_e^i, y_e^i, z_e^i, x_w^i, y_w^i, z_w^i, x_f^i, y_f^i, z_f^i)$ or angle-based features $V_a = (q_1^i, q_2^i, q_3^i, x_w^i, y_w^i, z_w^i, x_f^i, y_f^i, z_f^i)$

In respect to the automatic signature verifier (ASV) configuration, we use two different state-of-the-art ASVs. The aim is to test the versatility of the use of anthropomorphic features in two completely different ASVs, which are each based on different methodologies. This makes them suitable for covering a wide range of signature properties, as needed for a thorough assessment of the anthropomorphic features in terms of performance achieved.

Verifier 1). Function-based + DTW. Enrolled and questioned signatures are compared by using the same DTW configuration proposed in [6]. The feature vector can be built using position-based features, $V_p = ((x_e^i, y_e^i, z_e^i, x_w^i, y_w^i, z_w^i, x_f^i, y_f^i, z_f^i)$ or angle-based features, $V_a = ((q_1^i, q_2^i, q_3^i, q_4^i, q_5^i, q_6^i)$ .Next, the first and second order time derivatives are added to the feature vector and, finally, a z-score normalization along the index i is performed.

Verifier 2). Histogram-based + Manhattan distance. The feature vector consists of two histograms with absolute and relative frequencies. The similarity between the reference and questioned features is then obtained from the Manhattan distance [8].

This verifier builds the histograms by using the pen-tip trajectory x; y converted into polar coordinates. As such, a 3D geometrical projection of the VSA model, positioned on a 2D plane, was performed through a normal vector. The normal vector n´ is deduced from the signature in the 3D plane: $P_\tau(x,y,z)$ $\xrightarrow{n'}$ $P_\tau(x,y)$

In our case, $\tau$ refers to the three considered histograms, which are related to the position-based anthropomorphic features. Specifically, they are the elbow ($h_e$), the wrist ($h_w$) and the finger ($h_f$). The histogram extraction of each joint position is carried out similarly to the original verifier [8]. Finally, we develop the histogram vector by concatenating the individual vectors of each joint position: $h_p = [\ h_e \| h_w \| h_f]$.

This verifier was also adapted for the case of the anglebased anthropomorphic features, $Q(q_k^i)$ of a signature. Let $\Delta V_a^i$ and $\Delta\Delta V_a^i$ be the first and second order differences between adjacent elements of $V_a^i$ , which are then used to process three kinds of histograms. The first type is a one dimensional histogram. It is related to relative frequency of attributes from $\Delta V_a^i$. Let $\forall k \in (1, \ldots, 6)$, $\Delta q_k^i$ be a relative attribute so, in a way similar to [8], a histogram of sixteen bins is processed. Because the range of the six joint angles is reduced compared with the angles used in the original algorithm, the minimum and maximum of the histogram were set at $[\mu_{\Delta q_k^i} \pm d \cdot \sigma_{\Delta q_k^i}]$ where d = 2, which comprises 95 percent of the values of $\Delta q_k^i$. As such, we obtain the histograms, $\forall k \in (1, \ldots, 6), h_{\Delta q_k^i}$. We concatenate them to obtain the first type of histograms: $h_1 = [h_{\Delta q_1^i} || \ldots || h_{\Delta q_6^i}]$. For the second type, we process the $\Delta \Delta V_a^i$ features in a way similar to the previous type. As in the original proposal, we use 24 bins to define the histogram $h_2$. The third type is a two dimensional histogram ($h_3$), which is related to the three consecutive coordinates of the angles. The input attributes for this histogram: $\forall k \in (1, \ldots, 6)$, $i \in (1, \ldots, m)$, $\{\Delta q_k^{i=1}, \ldots, \Delta q_k^{i=m-1}, \Delta q_k^{i=1}, \ldots, \Delta q_k^{i=m-2}\}$, $\{\Delta q_k^{i=2}, \ldots, \Delta q_k^{i=m}, \Delta q_k^{i=3}, \ldots, \Delta q_k^{i=m}\}$.. Finally, the three types of angle-based histograms are concatenated: : $h_a = [h_1 || h_2 || h_3]$.

It should be noted that in our implementation, the absolute and relative epsilons were experimentally fixed at 0.4 and 0.004.

Combination. Among the different approaches for combining biometric systems, in this paper, we propose to fuse the position-based and angle-based anthropomorphic features in a single ASV, at feature and score level. In the case of feature level fusion, the feature matrix is obtained as: $V = [V_p, \Delta V_P, \Delta \Delta V_P, V_a, \Delta V_a, \Delta \Delta V_a]$ in the DTW-based ASV. In the Manhattan-based ASV, the histogram vector of each signature is built as follows: $h_p = [h_p || h_a]$. In the case of score level fusion [19], this is performed via a weighted sum: $s_f = \omega \cdot s_p + (1 - \omega) \cdot s_a$, where $0 < \omega < 1$ is the weighting, $s_f$ the final score, $s_p$ the score using only position-based features in the ASV and $s_a$ using only anglebased features. Furthermore, a tanh-estimator was used to normalize the scores $s_p$ and $s_a$ in the range (0,1).

TABLE 3
Influence of Pen-Tip Angles

| Signature | $(\theta_r^i, \phi_r^i)$ | | $(\theta_s^i, \phi_s^i)$ | | $(\theta_p^i, \phi_p^i)$ | | $(\theta_f^i, \phi_f^i)$ | |
|---|---|---|---|---|---|---|---|---|
| Scale | RF | SF | RF | SF | RF | SF | RF | SF |
| 1:10 | 32.97 | 14.92 | 27.57 | 14.60 | 0.85 | 4.64 | 0.75 | 3.52 |
| 1:1 | 24.14 | 12.28 | 16.49 | 10.24 | 0.75 | 4.64 | 0.75 | 3.44 |
| 10:1 | 6.83 | 7.52 | 4.87 | 6.96 | 1.00 | 5.24 | 0.90 | 5.68 |

*underline denotes the best achieved performance in SF.*
*Results in EER (%).*

# 5. INITIAL SETUP CONFIGURATION OF THE VIRTUAL SKELETAL ARM MODEL

In this section, we attempt to identify the best performing configuration of the virtual skeletal arm model for signature verification. Specifically, we seek the best pen-tip orientation for the VSA, how best to manage the transitions between pendown/pen-up and the best initial posture of the VSA. All experiments in this section are conducted with the MCYT-100 corpus [20] and the DTW [6]. The feature matrix required for the DTW is built using only the angle-based anthropomorphic features: $V = [V_a, \Delta V_a, \Delta \Delta V_a]$. For repeatability, only the first five signatures are used for training and the remaining genuine signatures for testing and building the FRR curve. To calculate the FAR curve, we conduct two experiments: one with random forgeries by using the first genuine signature of all users and the other with skilled forgeries, using all available skilled forgery signatures in the corpus. The Equal Error Rate (EER) is used to measure the final performance.

## 5.1 Configuration of the Pen-Tip Orientation

To calculate the angle-based anthropomorphic features, the CF {$S_6$} needs to be correctly positioned and orientated. In order to achieve this, the 3D position of the signature ; $(p_y^i, p_x^i, p_z^i)$ is used, where $p_z^i$ a binary vector set to the penup (+5 mm) and pen-down (0 mm) transitions and the pentip orientation with the angles $(\theta^i, \phi^i)$.

To choose the optimal setting of pen-tip angles, we study four possible configurations. First, real azimuth and inclination angles ($\theta_r^i, \phi_r^i$) are used, since MCYT-100 provides them. Second, these angles are smoothed ($\theta_s^i, \phi_s^i$) with a moving average filter and 15 samples in the span interval. Third, because azimuth and inclination might not be available, these angles are estimated ($\theta_p^i, \phi_p^i$), using the same configuration as the ink deposition model proposed in [21]. Fourth, the angles are fixed in a neutral position, i.e., $\theta_f^i = \pi/3$ and $\phi_f^i = 3\pi/4$.

The initial configuration of the VSA in this experiment mimics a standard position for writing. As such, the initial six joints' angles are set to: $Q(q_k^{i=0}) = (0, \frac{3\pi}{4}, \frac{-2\pi}{3}, 0, \frac{\pi}{2}, 0)$. Consequently, the initial pen position is calculated through the forward kinematics (Section 3.1), thus obtaining: $^0 p_6^{i=0} = [475.29, 0, -73.65]^T mm$.

The use of pen-tip angles in signature verification is a matter for debate. Some authors suggest that using them as function-based features is beneficial for the performance [22]. Other authors found a deterioration in the final results [23].

We do not use them directly as features in a ASV system, but instead we use them for orientating the CF $\{S_6\}$.

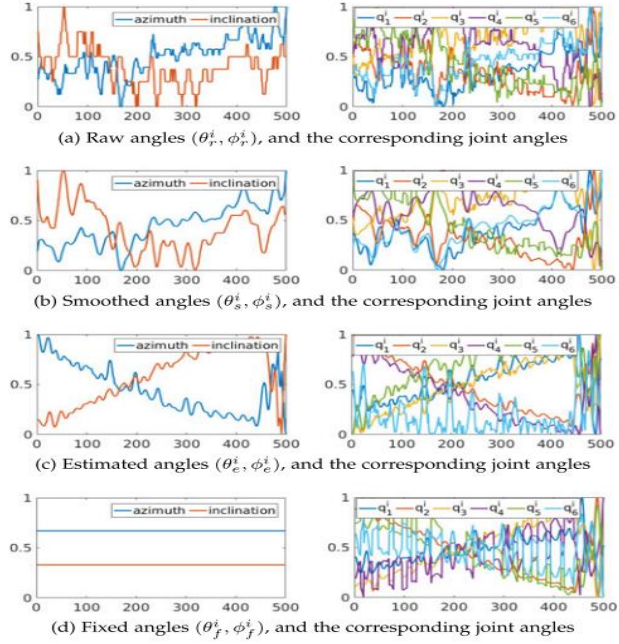

(a) Raw angles ($\theta_r^t$, $\phi_r^t$), and the corresponding joint angles

(b) Smoothed angles ($\theta_s^t$, $\phi_s^t$), and the corresponding joint angles

(c) Estimated angles ($\theta_e^t$, $\phi_e^t$), and the corresponding joint angles

(d) Fixed angles ($\theta_f^t$, $\phi_f^t$), and the corresponding joint angles

Fig. 7. Example of the effects of different pen-tip angles on the orientation of the CF $\{S_6\}$ in the angle-based anthropomorphic features.

The results, as set out in Table 3, suggest that the real acquired pen-tip angles (both in their raw form or smoothed) introduce confusion in the calculation of the six joint angles. The main reason for this is probably anomalies in the quantification of the pen-tip angles in the device used. Associated errors are also propagated when they are smoothed. As an example, Figs. 7a and 7b show the effect of using the raw and smoothed angles from a particular signature from MCYT-100 on the angle-based anthropomorphic features obtained. We can observe such quantification in the provided pen-tip angles as well as the confusion in the obtained angle-based features. This seems to validate the idea that the provided pen-tip angles are not sufficiently accurate for our purposes. Better results were obtained by using the estimated pair of pen-tip angles from the trajectory. Thus we can gain a degree of smoothness in the pen-tip angles and better definition of angle-based signatures, as shown Fig. 7c. However, the best results are found when they are fixed in a definite ergonomic posture. In the Fig. 7d, on the right, we can observe clearly abrupt transitions in the obtained angle-based features when pen-tip angles are fixed. They correspond to the pen-down and pen-up transitions and seem to be helpful to the operation of the signature verifier.

TABLE 4
Pen-Down & Pen-Ups Effect

| Signature Scale | $p_d$: $p_z^t = 0$ $p_u$: $p_z^t = +5$ | | $p_d$: $p_z^t = 0$ $p_u$: $p_z^t = 0$ | | $p_d$: $p_z^t = 0$ $p_u$: $p_z^t = 0$, $q_6^t + 1°$ | |
|---|---|---|---|---|---|---|
| | RF | SF | RF | SF | RF | SF |
| 1:10 | 0.75 | 3.52 | 0.90 | 3.56 | 1.09 | 4.08 |
| 1:1 | 0.75 | 3.44 | 0.94 | 3.48 | 0.80 | 3.52 |
| 10:1 | 0.95 | 5.68 | 1.19 | 5.72 | 1.13 | 5.68 |

*underline* denotes the best achieved performance in SF.
*Results in ERR (%).*

Furthermore, the influence of scaling the signature is also examined. The performance is calculated with the signature trajectory in millimeters (same unit of length as in the VSA model, i.e., 1:1), ten times smaller (1:10) and ten times bigger (10:1). It is observed in Table 3 that incrementing or decreasing the pen-tip trajectory ten times did not produce better results. In addition to being the best result, and for simplicity, we use fixed angles and coordinates without scaling. This configuration can also be used when the pen-tip angles are not given, as in the majority of publicly available, on-line signature databases.

## 5.2 Pen-Up Lift Influence on the VSA Model

Having fixed the pen-tip angles at $\theta_f^i$ =π/3 and $\emptyset_f^i$ = 3π/4., and to obtain better performance, we study how the VSA should manage the pen lift during the transitions from pen-down $p_d$ to pen-up $p_u$. For this purpose, three possible configurations were studied. First, we use $p_z^i$ =0 for pen-downs and $p_z^i$ =5 mm, for pen-ups, which simulates an average pen-up height of five millimetres. Next, we put $p_z^i$ = 0 for all signatures. Finally, we again set $p_z^i$ = 0 and increment the last joint angle $q_6^i$ by one degree during the pen-ups in order to mimic the twist of the wrist in the pen-ups.

Experimental results are shown in Table 4. In general terms, all configurations result in a similar performance for each pen-tip modality. This suggests that the pen-lift information helps the system reach the correct decision in the classifier. Finally, and once again, the best results are obtained when we set the pen-tip trajectory scaling factor to 1.

## 5.3 Optimal VSA Initial Posture

It is possible for the VSA to write in a horizontal plane, as on a table, in a vertical plane, as on a wall, or on an inclined plane, as in writing on a lectern. In this section we identify the writing plane in which the angle-based anthropomorphic features provide the best results in an ASV. The considered writing planes are illustrated by the diagrams in Table 5. These planes are rotated in respect to Γ=($\rho_x, \rho_y, \rho_z$). Accordingly, we geometrically rotate the pen-tip position and angles ( $p_x^i$, $p_y^i$, $p_z^i$, $\theta_f^i$, $\emptyset_f^i$ ) order to guarantee a normal vector from Γ.

Table 5 shows the performance when the signature's initial point changes according to the horizontal $h$ and vertical $v$ levels, which are depicted in the figures included in Table 5. In general, we can see a quite stable performance in many of the studied starting points. For a quantitative analysis of the performance, we focus on skilled forgeries. The worst performance is seen when the VSA writes on the ceiling Γ= (0, -π, 0), which almost certainly coincides with the most uncomfortable position for a real signer. The best performance is observed in the horizontal position. This not only requires rotations in the pen-tip but also coincides with the usual plane of human signing.

### TABLE 5
### Influence of the Initial Posture of the VSA Model

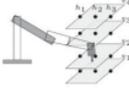

Γ = (0, 0, 0)  *like writing on a table*

|  | Random Forgery |  |  | Skilled Forgery |  |  |  |
|---|---|---|---|---|---|---|---|
|  | $h_1$ | $h_2$ | $h_3$ | $h_1$ | $h_2$ | $h_3$ |  |
|  | 0.80 | 0.75 | 0.75 | 4.00 | 4.16 | 3.64 | $v_4$ |
|  | 0.75 | 1.00 | 0.80 | 3.72 | 4.28 | 3.44 | $v_3$ |
|  | 3.24 | 0.85 | 0.85 | 7.08 | 4.00 | 3.48 | $v_2$ |
|  | 0.80 | 0.75 | 0.75 | 3.64 | 4.16 | 3.68 | $v_1$ |

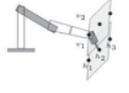

Γ = (0, -π/4, 0)  *like writing on a sloping table*

|  | Random Forgery |  |  | Skilled Forgery |  |  |  |
|---|---|---|---|---|---|---|---|
|  | $h_1$ | $h_2$ | $h_3$ | $h_1$ | $h_2$ | $h_3$ |  |
|  | 0.85 | 0.75 | 0.90 | 4.08 | 3.76 | 3.48 | $v_2$ |
|  | 2.29 | 0.75 | 1.85 | 6.92 | 3.68 | 5.60 | $v_1$ |

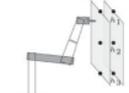

Γ = (0, -π/2, 0)  *like writing on a wall*

|  | Random Forgery |  |  | Skilled Forgery |  |  |  |
|---|---|---|---|---|---|---|---|
|  | $h_1$ | $h_2$ | $h_3$ | $h_1$ | $h_2$ | $h_3$ |  |
|  | 0.80 | 0.80 | 0.80 | 4.20 | 4.00 | 3.68 | $v_2$ |
|  | 4.03 | 0.80 | 0.90 | 7.32 | 3.88 | 4.00 | $v_1$ |

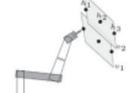

Γ = (0, -3π/4, 0)  *like writing on a sloping ceiling*

|  | Random Forgery |  |  | Skilled Forgery |  |  |  |
|---|---|---|---|---|---|---|---|
|  | $h_1$ | $h_2$ | $h_3$ | $h_1$ | $h_2$ | $h_3$ |  |
|  | 0.85 | 0.90 | 0.80 | 3.72 | 3.68 | 3.60 | $v_2$ |
|  | 0.80 | 0.80 | 1.00 | 4.20 | 3.64 | 4.04 | $v_1$ |

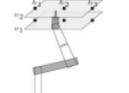

Γ = (0, -π, 0)  *like writing on a ceiling*

|  | Random Forgery |  |  | Skilled Forgery |  |  |  |
|---|---|---|---|---|---|---|---|
|  | $h_1$ | $h_2$ | $h_3$ | $h_1$ | $h_2$ | $h_3$ |  |
|  | 1.60 | 8.17 | 0.80 | 4.12 | 10.60 | 3.92 | $v_2$ |
|  | 2.70 | 4.44 | 0.80 | 8.60 | 6.92 | 3.88 | $v_1$ |

underline denotes the best achieved performance in SF.
*Performance in EER (%)*

Therefore, from Table 5, we conclude that the best initial posture corresponds to *Γ=(0,0,0), h₃ and v₃,* which is equivalent to $Q(q_k^{i=0}) = (0, \frac{3\pi}{4}, -\frac{2\pi}{3}, 0, \frac{\pi}{2}, 0)$, *and* $^0 p_6^{i=0} = [ 475.29, 0, -73.65 ]^T mm$ .

## 5.4 Weighting Parameter Adjustment for Fusing Anthropomorphic Features at Score Level

Finally, the weighting parameter, v, needs to be adjusted if score level fusion is used. As such, we obtain 0.85 and 3.88

percent for EER in RF and SF experiments by using the position-based features. Using the angle-based features, 0.75 and 3.44 percent EER was obtained for RF and SF, respectively. We can fuse them by changing the weighting parameter: $\omega \in \{0.0, 0.2, 0.4, 0.6, 0.8, 1.0\}$. The best results are obtained for $\omega = 0.4$, which are EER= 0.80% in RF and EER= 3.24% in SF.

We can see that both fusion methods maintain the results in RF and improve the performance in SF. We also see that both kinds of anthropomorphic features give a competitive performance when they are used independently. This may be a reason for the better performance in the score level combination being found at $\omega$ =0.4, which has a similar relevance for each score.

To sum up this section, the suggested configuration of the VSA model for generating the most effective anthropomorphic features is: 1) fixing pen-tip angles $\theta_f^i = \pi/3$ and $\phi_f^i = 3\pi/4$, which represent an ergonomically acceptable position; 2) a signature scale of 1:1 in millimeters, which corresponds to that used elsewhere in the VSA model; 3) $p_z^i$ set to zero during the pen-downs and five millimeters in the pen-ups; 4) the VSA configured to write on a horizontal table, i.e., $\Gamma$=(0,0,0); and 5) a similar weighting factor for the fusion at score level of the two kinds of anthropomorphic features, i.e., $\omega = 0.4$. We use this configuration in the remainder of this work.

TABLE 6
EER Results (in %) Using the Anthropomorphic Features
Generated by the Virtual Skeletal Arm Model Against
the Third Party Databases and Verifiers Used

| | MCYT-100 | | | | | | | |
|---|---|---|---|---|---|---|---|---|
| | Position | | Angles | | Feature level | | Score level | |
| | RF | SF | RF | SF | RF | SF | RF | SF |
| DTW | 0.85 | 3.88 | 0.75 | 3.44 | 0.76 | 3.32 | 0.80 | 3.24 |
| MAN | 1.60 | 4.92 | 3.34 | 7.36 | 1.55 | 4.68 | 1.57 | 4.52 |
| | MCYT-330 | | | | | | | |
| | Position | | Angles | | Feature level | | Score level | |
| | RF | SF | RF | SF | RF | SF | RF | SF |
| DTW | 0.86 | 3.53 | 0.77 | 3.92 | 0.79 | 3.60 | 0.77 | 3.48 |
| MAN | 1.97 | 5.50 | 3.48 | 7.64 | 1.82 | 5.20 | 1.81 | 5.02 |
| | BiosecurID-SONOF | | | | | | | |
| | Position | | Angles | | Feature level | | Score level | |
| | RF | SF | RF | SF | RF | SF | RF | SF |
| DTW | 1.38 | 2.34 | 1.45 | 2.40 | 1.34 | 2.21 | 1.47 | 2.27 |
| MAN | 2.27 | 3.91 | 3.65 | 4.36 | 2.18 | 3.47 | 2.28 | 3.41 |
| | SUSIG-Visual | | | | | | | |
| | Position | | Angles | | Feature level | | Score level | |
| | RF | SF | RF | SF | RF | SF | RF | SF |
| DTW | 0.98 | 3.30 | 0.71 | 5.32 | 0.85 | 3.72 | 0.78 | 3.83 |
| MAN | 4.44 | 9.04 | 7.30 | 9.26 | 4.23 | 8.83 | 4.18 | 8.83 |
| | SUSIG-Blind | | | | | | | |
| | Position | | Angles | | Feature level | | Score level | |
| | RF | SF | RF | SF | RF | SF | RF | SF |
| DTW | 0.26 | 2.84 | 0.12 | 3.41 | 0.26 | 2.95 | 0.09 | 2.95 |
| MAN | 1.84 | 6.02 | 3.50 | 6.25 | 1.84 | 5.68 | 1.57 | 5.57 |
| | mobile SG-NOTE | | | | | | | |
| | Position | | Angles | | Feature level | | Score level | |
| | RF | SF | RF | SF | RF | SF | RF | SF |
| DTW | 1.00 | - | 1.17 | - | 1.00 | - | 0.50 | - |
| MAN | 3.00 | - | 6.33 | - | 4.17 | - | 4.17 | - |
| | OnOffSigBengali-75 | | | | | | | |
| | Position | | Angles | | Feature level | | Score level | |
| | RF | SF | RF | SF | RF | SF | RF | SF |
| DTW | 0.86 | - | 0.56 | - | 0.61 | - | 0.68 | - |
| MAN | 4.83 | - | 6.81 | - | 4.41 | - | 4.43 | - |
| | OnOffSigDevanagari-75 | | | | | | | |
| | Position | | Angles | | Feature level | | Score level | |
| | RF | SF | RF | SF | RF | SF | RF | SF |
| DTW | 1.28 | - | 0.99 | - | 0.74 | - | 0.92 | - |
| MAN | 5.21 | - | 7.64 | - | 4.85 | - | 4.95 | - |

# 6. RESULTS

The experiments we conducted are aimed at assessing the usefulness of the anthropomorphic features in on-line ASV. For a fair validation, we tested the anthropomorphic features with a DTW-based verifier (DTW) and a Manhattan-based one (MAN). We also use multiple publicly available databases in different scripts and acquired using different devices. These are the third party databases: MCYT-100 and MCYT-300 [20], BiosecurID-SONOF [2], SUSIG-Visual and Blind [24], mobile SG-NOTE [25], OnOffSigBengali-75 and OnOffSigDevanagari-75 [26].

## 6.1 Performance Analysis

For the DTW verifier, the results are presented in Table 6. By using the MCYT-100 and MCYT-300 databases, promising results are found in all cases, which shows the validity of fusing the anthropomorphic features. Similar findings are observed with the BiosecurID-SONOF corpus, which confirm the hypothesis of the usefulness of anthropomorphic features in ASV. Once again, good results are found using the SUSIG database. The SUSIG-Visual corpus gives a better performance when the position-based anthropomorphic features are used compared with that with angle-based anthropomorphic features. But after fusion a more competitive performance was obtained. In the case of the SUSIG-Blind corpus, a promising improvement was achieved in RF after the fusion at score level. In addition, a mobile database was used for the purposes of this paper. Significant improvements were observed in the fusion level score, where the EER was reduced by more than a half. Finally, these features were tested using an on-line database with signatures in non-Western scripts. For the case of the Bengali script, we observe that the position features seem to be more efficient than the angle features. However, the final performance is still in the same range of values compared with that from the other databases. Similar findings are observed for the on-line Devanagari database, where we obtained the best performance after fusion. These results suggest that the anthropomorphic features are useful for signature verification in non-Western script scenarios.

For the MAN verifier, experimental results are also given in Table 6. As an overall perspective of this verifier, we observe that the RF experiment gives a better performance than SF in all cases. Again, in all cases, we can observe that the position-based features report slightly better performance compared with the angle-based features. However, the latter type of feature introduces information to the classifier after the fusion schemes for most databases used.

For the MCYT database, we can see a slightly better performance for the MCYT-100. However, the results were consistently improved after fusion. In the BiosecurID-SONOF corpus, the performance results after fusion are quite similar for both combination strategies. Very good results are obtained after the fusion for SUSIG databases. For the SUSIG-Visual corpus both RF and SF experiments are improved, despite the differences between performance with position and angle features. In the case of the SUSIG-Blind corpus, a similar performance is obtained between angle and position as well as between fusion at feature and score level. In the case of the mobile database, better performance is achieved with the position features. However, the final results are still competitive and not really damaged by the negative influence of the angle feature results. When this verifier is tested with other scripts, namely Bengali and Devanagari, we can observe the robustness of the fusion in both cases, which always improves the individual performance results.

## 6.2 Performance Analysis of VSA Model When Including Arm and Forearm Lengths

The humerus, ulna and radius lengths are different for each person. Bearing this in mind, we decided to analyze the performance in signature verification through a more realistic configuration of our VSA model, denoted as $VSA_r$. Since these lengths are not provided in the dataset we used, a hypothetical situation is assumed by estimating them through Gaussian distributions, previously studied in [27]. The humerus is represented by $L_2$ in our $VSA_r$ and its length is modified for each signer using the length distributions in mm [27] of $N$ (334:15.8) for male and $N$ (307; 15.9) for female. The distance $L_3$ was reduced to one millimeter, as an epsilon. As the radius and ulna are represented in a single link, we have used the radius length distribution [27] to modify $L_4$, which are $N(265; 15.4)$ and $N(238; 10.7)$ for male and female respectively. The gender was randomly assigned to the signers. We then carried out a similar benchmark to that of Section 6.1. Note that the fusion at score level was conducted in the same conditions as previous experiments, i.e., $\omega = 0.4$.

Experimental results are given in Table 7 and can be compared to the results in Table 6. In general terms, while we may not observe significant improvement in the performance when only position features are considered, significant improvements are observed when only angle features are used in the verifiers. This suggests that angle features are positively more useful when the realistic bone lengths are included in the VSA.

On random forgeries, the performance is similar in both experiments, as it can see in Tables 6 and 7. In some cases, we can observe positive improvement (e.g., EER = 2.28% versus EER = 1.86% in BiosecurID-SONOF with Manhattan after fusion at score level); similar results (e.g., EER = 0.79% in MCYT-330 after fusion at feature level with DTW) or some slight worsening (e.g., EER = 3.00% versus EER = 3.67% in mobile SG-NOTE by considering the position features in the Manhattan-based verifier).

On skilled forgeries, very good improvements are observed after fusion in all experiments, as we can observe by comparing Tables 6 and 7. This suggests that the upper arm bone lengths can be used as a soft biometric for signature verification, because of the significant improvements in the final performance results with a large number of databases and two completely different verifiers.

## 6.3 Comparative Results in Signature Verification

In Table 8 we show comparative results between our proposed VSA system and other competitive state-of-the-art systems for on-line signature verification, ranked by their performance in terms of EER for SFs. The other systems shown used five signatures to train the system. While some systems have randomly selected these five signatures, in ours we used the first five signatures. This protocol leads to greater repeatability as well as to less intra-personal variability. Moreover, we have neither adapted nor fine-tuned our system for any dataset. We have used the same configuration in our proposed system with all databases. Despite not reporting the best performance in all cases, for the sake of uniformity, the related entry in the table of our system corresponds to fusion at score level in the DTW-based system

On the MCYT-100, the best performance was achieved by [12]. It is worth pointing out that only two signature datasets were evaluated in [12] and [7]. Additionally, the authors reported the best results after fine-tuning three parameters in their system in [12] and one parameter in the system of [7]. Moreover, neither [12] nor [7] presented results with the RF experiments. Instead, we apply the same configuration of our system to all datasets. This, in our opinion, might lead to a more realistic scenario, where a system is designed for a particular dataset and blind tested with other corpuses.

On the MCYT-330, very promising improvements on the competitive state-of-the-art are manifested with our proposal. To the best of our knowledge, this dataset has not been used frequently in recent works, but MCYT-100 has and seems to be more popular in the community at the moment.

**TABLE 7**
**EER Results (in %) Using the Anthropomorphic Features Generated by the Realistic Virtual Skeletal Arm (VSAr) Model Against the Third Party Databases and Verifiers Used**

| | Position | | Angles | | Feature level | | Score level | |
|---|---|---|---|---|---|---|---|---|
| | RF | SF | RF | SF | RF | SF | RF | SF |
| **MCYT-100** | | | | | | | | |
| DTW | 0.85 | 3.80 | 0.80 | 2.52 | 0.75 | 2.64 | 0.75 | 2.68 |
| MAN | 1.29 | 4.32 | 3.44 | 6.52 | 1.34 | 3.84 | 1.55 | 4.24 |
| **MCYT-330** | | | | | | | | |
| DTW | 0.97 | 3.99 | 0.74 | 2.86 | 0.79 | 2.76 | 0.74 | 3.09 |
| MAN | 1.67 | 4.53 | 3.41 | 7.14 | 1.58 | 4.34 | 1.65 | 4.57 |
| **BiosecurID-SONOF** | | | | | | | | |
| DTW | 1.38 | 2.27 | 1.38 | 2.08 | 1.38 | 1.89 | 1.31 | 1.89 |
| MAN | 2.20 | 3.60 | 4.06 | 4.23 | 2.15 | 3.16 | 1.86 | 3.16 |
| **SUSIG-Visual** | | | | | | | | |
| DTW | 1.02 | 3.30 | 0.71 | 4.04 | 0.78 | 3.09 | 0.78 | 3.09 |
| MAN | 4.39 | 8.09 | 6.93 | 7.77 | 4.03 | 7.87 | 4.11 | 7.34 |
| **SUSIG-Blind** | | | | | | | | |
| DTW | 0.26 | 2.84 | 0.07 | 2.05 | 0.07 | 2.05 | 0.07 | 2.05 |
| MAN | 1.76 | 5.11 | 3.03 | 6.02 | 1.57 | 4.66 | 1.53 | 4.20 |
| **mobile SG-NOTE** | | | | | | | | |
| DTW | 1.00 | - | 1.17 | - | 1.00 | - | 0.50 | - |
| MAN | 3.67 | - | 5.50 | - | 4.33 | - | 4.17 | - |
| **OnOffSigBengali-75** | | | | | | | | |
| DTW | 0.90 | - | 0.63 | - | 0.65 | - | 0.63 | - |
| MAN | 4.00 | - | 5.96 | - | 3.42 | - | 3.93 | - |
| **OnOffSigDevanagari-75** | | | | | | | | |
| DTW | 1.28 | - | 0.70 | - | 0.41 | - | 0.77 | - |
| MAN | 4.34 | - | 6.70 | - | 4.13 | - | 3.86 | - |

**TABLE 8**
**On-Line Automatic Signature Verification Results in the State-of-the-Art**

| Database | Year | Method | RF | SF |
|---|---|---|---|---|
| MCYT-100 [20] | 2016 | DTW+VQ [12] | - | 1.55 |
| | 2018 | VSAr + DTW | 0.75 | 2.68 |
| | 2017 | WP+BL DTW fusion [7] | - | 2.76 |
| | 2018 | VSA + DTW | 0.80 | 3.24 |
| | 2018 | GMM+DTW [28] | - | 3.05 |
| | 2017 | ΣΛ + DTW [14] | 1.01 | 3.56 |
| | 2014 | Histogram+Manhattan [8] | 1.15 | 4.02 |
| | 2017 | Symbolic Rep [29] | 2.40 | 5.70 |
| MCYT-330 [20] | 2018 | VSAr + DTW | 0.74 | 3.09 |
| | 2018 | VSA + DTW | 0.77 | 3.48 |
| | 2015 | Two Stage DTW [6] | 1.06 | 3.94 |
| | 2010 | HMM [30] | - | 6.33 |
| | 2009 | Symbolic Representation [31] | 2.10 | 6.45 |
| | 2017 | Symbolic Representation [29] | 3.60 | 6.90 |
| BiosecurID-SONOF [2] | 2018 | VSAr + DTW | 1.31 | 1.89 |
| | 2017 | Histogram+Manhattan [5] | 1.16 | 1.98 |
| | 2018 | VSA + DTW | 1.47 | 2.27 |
| | 2017 | Function-based+DTW [5] | 0.23 | 3.08 |
| | 2015 | Fusion (Online+Synt. Offline) [2] | 0.63 | 5.09 |
| SUSIG-Visual [24] | 2009 | DTW+Linear Classifier [24] | 4.08 | 2.10 |
| | 2017 | ΣΛ + DTW [14] | 1.48 | 2.23 |
| | 2018 | VSA + DTW | 0.78 | 3.09 |
| | 2015 | Two Stage DTW [6] | 1.34 | 3.09 |
| | 2018 | VSAr + DTW | 0.78 | 3.83 |
| | 2014 | Histogram+Manhattan [8] | 2.91 | 4.37 |
| | 2014 | Fuzzy modelling [32] | 4.57 | 5.38 |
| SUSIG-Blind [24] | 2018 | VSAr + DTW | 0.07 | 2.05 |
| | 2017 | ΣΛ + DTW [14] | 0.79 | 2.05 |
| | 2009 | DTW+Linear Classifier [24] | 2.82 | 2.85 |
| | 2018 | VSA + DTW | 0.09 | 2.95 |
| mobile SG-NOTE [25] | 2018 | VSAr + DTW | 0.50 | - |
| | 2018 | VSA + DTW | 0.50 | - |
| | 2017 | ΣΛ + DTW [14] | 0.78 | - |
| | 2014 | Global Features+Mahalanobis [25] | 2.10 | - |
| OnOffSigBen-gali-75 [26] | 2017 | Function-based + DTW [26] | 0.27 | - |
| | 2018 | VSAr + DTW | 0.63 | - |
| | 2018 | VSA + DTW | 0.98 | - |
| OnOffSigDeva-nagari-75 [26] | 2017 | Function-based + DTW [26] | 0.31 | - |
| | 2018 | VSAr + DTW | 0.41 | - |
| | 2018 | VSA + DTW | 0.74 | - |

*Performance in EER (%).*

On the BiosecurID-SONOF [2], once again, our system gave the best results in SF, the RF results being improved by other systems [2], [5]. For this dataset there is little consistency in improvements for RF and SF.

On the SUSIG-Visual, the best results were found in 2009 [24], where only this corpus and SUSIG-Blind were tested. Our results are on par with [6] in the case of SF, and a clear improvement on all others in RF.

On the SUSIG-Blind, significant improvements were obtained by using the anthropomorphic features. Although the results are on par with [14] in SF, clear improvements can be observed throughout in RF. Compared with SUSIG-Visual, this corpus is not as popular as MCYT 330. However, SUSIG corpuses do contain unreal genuine signatures. This makes it difficult to correctly model the intra-personal variability for the systems.

On the mobile SG-NOTE, our system clearly outperforms the state-of-the-art. As this corpus does not include forgeries, there are no SF results. This dataset was collected with a Smartphone. This suggests that our system is also competitive with signatures collected with this kind of device.

On OnOffSigBengali-75, our system is still competitive with the current state-of-the-art. Our system was optimized by a particular Western database, i.e., MCYT-100. However, comparable performance results are obtained by testing signatures in other scripts.

On OnOffSigDevanagari-75, the obtained performance with our system is close to the first entry in the table. Note that the work presented in [26] only considered these two corpuses in the Indian language. Our performance indicates the benefits of using our proposed method in non-Western scripts such as Bengali or Devanagari.

# 7. CONCLUSIONS

In this paper, a novel set of anthropomorphic features for signature verification is introduced. These features are generated using the pen-tip position and orientation when signing on a digital tablet. To develop the anthropomorphic features, a Virtual Skeletal Arm model has been designed. For the sake of verifiability, this model is based

We find that the estimation of anthropomorphic features can be divided into two sets, according to how the motion of the arm is described: one as a sequence of 3D Cartesian positions of the arm joints and the other as a sequence of arm joint angles. These are defined as position-based and angle-based anthropomorphic features, respectively. We also highlight that the open source code for generating the anthropomorphic features can be freely downloaded from: www.gpds.ulpgc.es.

Based on these new anthropomorphic features, we have proposed how to exploit them in automatic signature verification. As such, two different state-of-the-art verifiers in terms of feature processing and classifying were used: one is based on function-based features following a DTW and the other is based on a histogram distribution of the features and a Manhattan distance classifier.

In addition, we have studied the capabilities of our VSA by incorporating the variations in the length of the arm bones. Although it may be impractical for application in the short-term, relevant improvements were observed as a proof of concept.

The proposed anthropomorphic features have outperformed the state-of-the-art performance on a wide variety of benchmark databases. It is worth noticing that experimental results were tested in challenging scenarios of multiple databases collected on several different digital devices (such as a Wacom Tablet or smartphones) and several languages and scripts (such as Western, Bengali and Devanagari).

The fact remains that a complete proposal for a signature verification system capable of considerably improving the results using any third party database remains an open challenge. In the meantime, our proposal seems to work efficiently and leads us to use additional human-like features in the signature verification task.